\documentclass[runningheads]{llncs}
\usepackage[T1]{fontenc}
\usepackage{hyperref}
\usepackage{color}

\usepackage[pdftex]{graphicx}
               
\usepackage{subfig}                 
\usepackage{amsmath}
\usepackage{amssymb}
\usepackage{booktabs}
\usepackage[misc]{ifsym}
\newcommand{\etal}{{et al.}}

\begin{document}
\title{Estimating Treatment Effects Under Heterogeneous Interference}
%
%
\author{Xiaofeng Lin\Letter
\and
Guoxi Zhang
\and
Xiaotian Lu
\and
Han Bao
\and
Koh Takeuchi
 \and
Hisashi Kashima
}
\authorrunning{X. Lin et al.}
\toctitle{Estimating Treatment Effects Under Heterogeneous Interference}
\tocauthor{Xiaofeng Lin, Guoxi Zhang, Xiaotian Lu, Han Bao, Koh
Takeuchi, Hisashi Kashima}

\institute{Graduate School of Informatics, Kyoto University, Kyoto, Japan
\\
\email{\{lxf,guoxi,lu\}@ml.ist.i.kyoto-u.ac.jp}; \email{\{bao,takeuchi,kashima\}@i.kyoto-u.ac.jp}
}

\maketitle
\sloppy
\begin{abstract}
Treatment effect estimation can assist in effective decision-making in e-commerce, medicine, and education. One popular application of this estimation lies in the prediction of the impact of a treatment (e.g., a promotion) on an outcome (e.g., sales) of a particular unit (e.g., an item), known as the individual treatment effect (ITE).  In many online applications, the outcome of a unit can be affected by the treatments of other units, as units are often associated, which is referred to as interference. For example, on an online shopping website,  sales of an item will be influenced by an advertisement of its co-purchased item. Prior studies have attempted to model interference to estimate the ITE accurately, but they often assume a homogeneous interference, i.e., relationships between units only have a single view. However, in real-world applications, interference may be heterogeneous, with multi-view relationships. For instance,  the sale of an item is usually affected by the treatment of its co-purchased and co-viewed items. We hypothesize that ITE estimation will be inaccurate if this heterogeneous interference is not properly modeled. Therefore, we propose a novel approach to model heterogeneous interference by developing a new architecture to aggregate information from diverse neighbors. Our proposed method contains graph neural networks that aggregate same-view information, a mechanism that aggregates information from different views, and attention mechanisms. In our experiments on multiple datasets with heterogeneous interference, the proposed method significantly outperforms existing methods for ITE estimation, confirming the importance of modeling heterogeneous interference.

\end{abstract}


\section{Introduction}

\begin{figure}[htbp]
	\centering
	\subfloat[Interference on a homogeneous graph]{
 \includegraphics[width=.45\columnwidth]{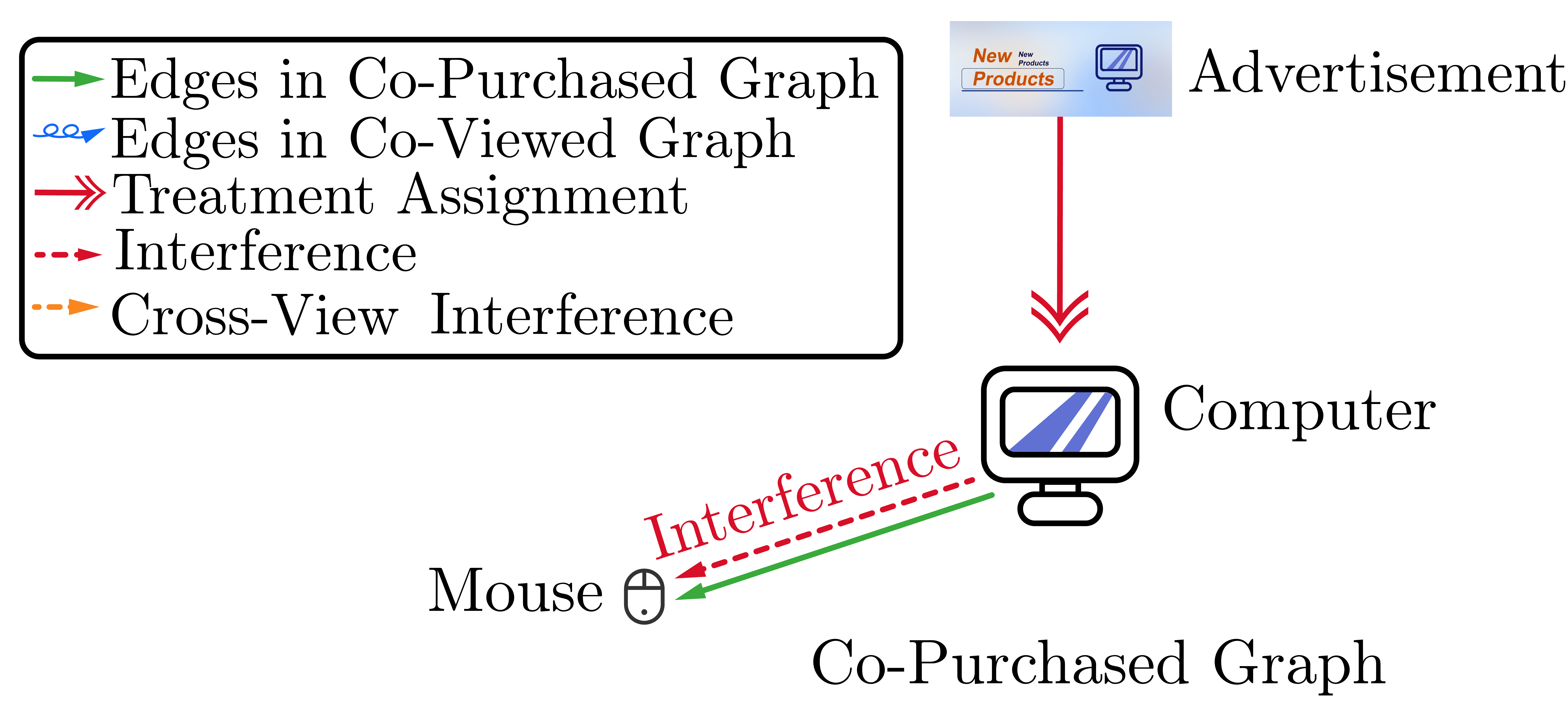}
 \label{fig:Interferencea}
    }
    \hspace{3pt}
	\subfloat[Interference on heterogeneous graphs]{
 \includegraphics[width=.45\columnwidth]{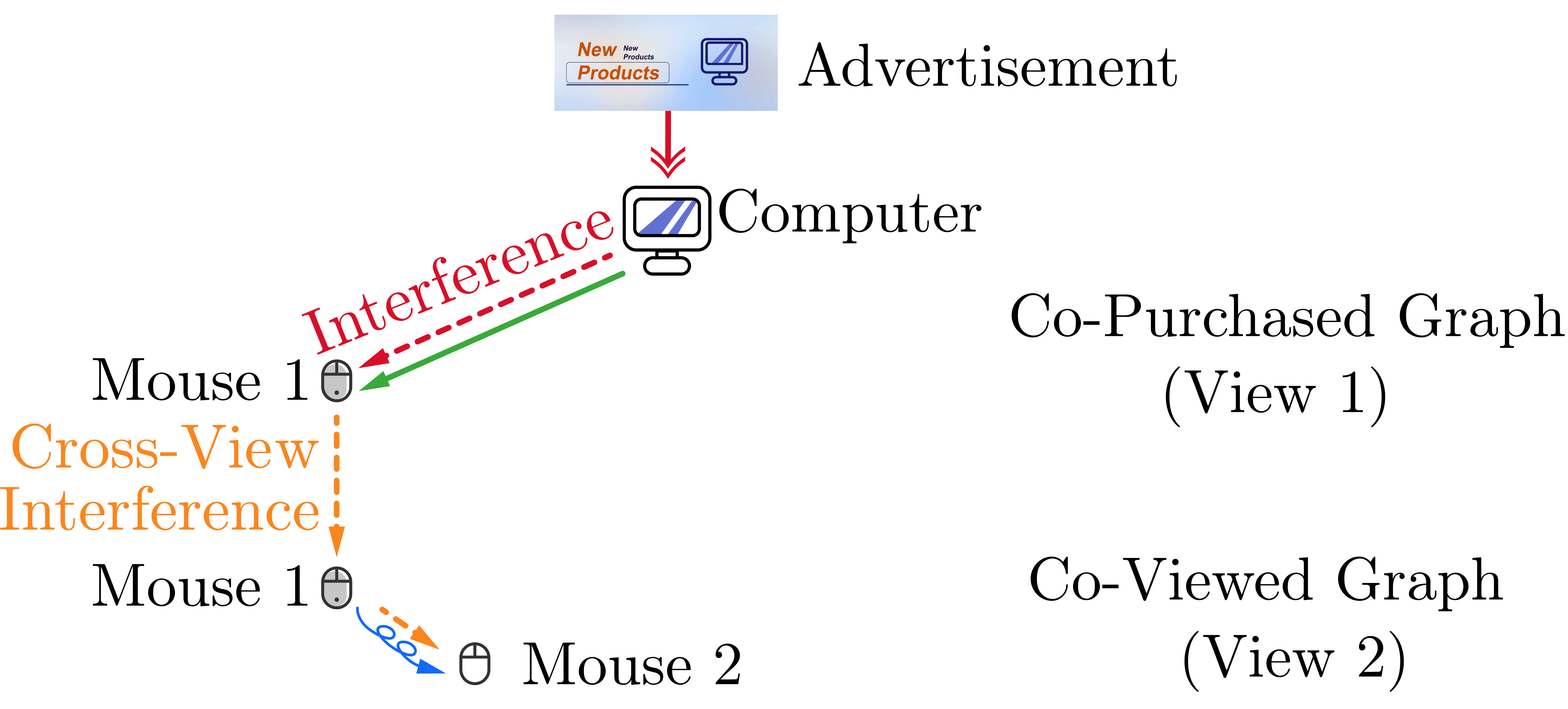}
 \label{fig:Interferenceb}
 }
\caption{An example of the difference between interference on a homogeneous graph and heterogeneous graphs. 
An edge in a co-purchased graph represents the relationship that both items are bought together by many customers, while an edge in a co-viewed graph represents the relationship that both items are viewed on an e-commerce platform together by many customers. Edges in different views or graphs constitute multi-view or heterogeneous edges.
}
\label{fig:Interference}
\end{figure}
In recent years, treatment effect estimation has been performed to enable effective decision-making in many fields, such as medicine~\cite{schnitzer2022estimands}, education~\cite{raudenbush2020randomized}, and e-commerce~\cite{nabi2022causal,sun2015causal,wang2015robust}. 
For example,  estimating treatment effects helps us understand whether an advertisement affects the sales of the advertised products.  The effect of a treatment (e.g., 
 advertisement) for a particular unit (e.g.,  product) is known as the individual treatment effect (ITE)~\cite{yao2021survey}, while that for a given group is known as the average treatment effect (ATE)~\cite{yao2021survey}.

This study aims to estimate treatment effects from observational graph data, which contain records of covariates of units,  relationships between units (i.e., graph structure), and treatment assignments with their outcomes. For example, data from an e-commerce platform typically include the logs of information regarding assignments of advertisements,  sales of items, item profiles, and relationships between items, e.g., a co-purchased relationship. 

As units are associated in these graphs, the outcome for a unit will be influenced by the treatments assigned to its neighboring units. This phenomenon is referred to as \emph{interference}~\cite{pmlr-v130-ma21c,rakesh2018linked}, an example of which is shown in Figure~\ref{fig:Interferencea}. In a co-purchased graph, many customers buy the Mouse when they buy the Computer. In this case, advertising the Computer may also influence the sales of the Mouse, whose sales can no longer be independent of the advertisement, making it challenging to estimate the ITE accurately.
Previous works have attempted to accurately estimate ITE given graph data by modeling interference, such as \emph{group-level interference}~\cite{doi:10.1198/016214508000000292,liu2014large,tchetgen2012causal}, which is a \emph{partial interference} and models interference within subgroups of units but ignores inter-group interference; \emph{pairwise interference}~\cite{Aronow2017EstimatingAC,doi:10.1080/01621459.2020.1768100,rakesh2018linked,RePEc:arx:papers:1906.10258}, which considers interference from  immediate neighbors only; and \emph{networked interference}~\cite{pmlr-v130-ma21c}, which can model interference from distant neighbors. All these methods assume single-view interference, such that a graph is homogeneous and can only represent the same relationship among units, such as a co-purchased graph. 

However, real-world graphs are rarely homogeneous, e.g., YouTube dataset~\cite{tang2009uncoverning}, and Amazon dataset~\cite{he2016ups}.
Therefore, we consider addressing interference on heterogeneous graphs that have multi-view edges,  such as co-viewed and co-purchased item-to-item graphs of the Amazon dataset~\cite{he2016ups}. 
In this case, units are influenced by treatments of their heterogeneous neighbors via the multi-view edges, which is referred to as \emph{heterogeneous interference} and often leads to \emph{cross-view interference}, an example of which is shown in Figure~\ref{fig:Interferenceb}.  Although there is no direct edge between the Computer and the Mouse 2, the advertisement of the Computer still affects sales of the Mouse 2 via the edge between the Computer and the Mouse 1 in the co-purchased graph and the edge between the Mouse 1 and the Mouse 2 in the co-viewed graph. Without properly modeling the heterogeneous interference, the cross-view interference cannot be addressed, which will result in inaccurate ITE estimation.

To overcome the difficulty caused by heterogeneous interference, we propose a novel method called \textbf{I}ndividual \textbf{T}reatment \textbf{E}ffects Estimator Under \textbf{H}eterogeneous \textbf{In}terference (HINITE; see Figure~\ref{fig:HINITE}).
The core idea of HINITE is to model the propagation of heterogeneous interference across units and views.
To this end, inspired by Wang~\etal~\cite{wang2019heterogeneous}, we design a heterogeneous information aggregation (HIA) layer, as shown in Figure~\ref{fig:hia}. 
In the HIA layer, multiple single-layered graph neural networks (GNNs)~\cite{kipf2016semi} are used to capture information within the same views, and a view-level information aggregation mechanism is then used to combine information from different views.
To properly model heterogeneous interference, the HIA layer also infers importances of different edges and views of heterogeneous graphs by applying attention mechanisms~\cite{vaswani2017attention,vel2018graph,wang2019heterogeneous}.
A single HIA layer can help units aggregate information from their 1-hop or direct neighbors across all views of heterogeneous graphs, enabling the HINITE to model the propagation of cross-view interference  by  stacking multiple HIA layers. Other components of the HINITE are explained in Section~\ref{sec:proposed}.

The contributions of this study can be summarized as follows:
\begin{itemize}
    \item This study describes a new issue of interference on heterogeneous (multi-view) graphs. Moreover, we formalize the problem of estimating ITE under heterogeneous interference.
    \item This study proposes a method to address interference on heterogeneous graphs with multi-view edges.
    \item Results of extensive experiments reveal that the proposed method outperforms existing methods for estimating ITE under heterogeneous interference while confirming the importance of modeling heterogeneous interference.
\end{itemize}

\section{Problem setting}
In this study, we aim to estimate ITE from observational heterogeneous graphs. Herein, we use $\boldsymbol{x}_i \in \mathbb{R}^d$ to denote the covariates of a unit $i$ (e.g., brand), $t_i \in \{0,1\}$ to denote the treatment assigned to a  unit $i$ (e.g., an advertisement), $y_{i} \in \mathbb{R}$ to denote the observed outcome of a unit $i$ (e.g., the observed sales of a unit $i$), and non-bold, italicized, and capitalized letters (e.g., $X_i$) to denote random variables. Moreover, a unit with $t=1$ is treated, and $t=0$ is controlled.
\paragraph{Homogeneous graphs.} Homogeneous graphs have only a single view of edges.   We use an  adjacency matrix $\bold{A} \in \{0,1\}^{n \times n}$ to represent the structure of a homogeneous graph, where $n$ is the number of nodes (units). If there is an edge between units $j$ and $i$, $A_{ij}$ = 1; otherwise, $A_{ij} = 0$. We let $A_{ii} = 0$.

\paragraph{Heterogeneous graphs.} This study considers heterogeneous graphs\footnote{Heterogeneous graphs can be classified into two types: those with multiple types of nodes and multiple types (views) of edges~\cite{Tang-etal09-ICDM}, and those with a single type of node and multiple types of edges~\cite{Tang-etal09-ICDM}. In this study, we focus on the latter type.} that have multiple views of edges~\cite{Tang-etal09-ICDM}, which are called heterogeneous or multi-view edges. We use the $\bold{H} = \{\bold{A}^{v}\}_{v=1}^{m}$ to denote all the multi-view graph structures, where $\bold{A}^{v}\in \{0,1\}^{n \times n}$ denotes the adjacency matrix of the $v$-th view, and $m$ is the number of views. We use $\bold{N}_{i}^v$ to denote the set of neighboring units of the unit $i$ in the $v$-th view,  $\bold{N}_{i}=\{\bold{N}_{i}^v\}^m_{v=1}$ to denote the set of neighbors of the unit $i$ across all views. Here, the units in $\bold{N}_{i}$ are heterogeneous neighbors of the unit $i$. 

\paragraph{ITE estimation without interference.} In traditional treatment effect estimation~\cite{rubin1980randomization,yao2021survey}, non-graph data are given and it is assumed that there is no interference between units~\cite{rubin1980randomization,yao2021survey}. In this case, the potential outcomes $y_{i}^{1}$ and $y_{i}^{0}$ of a unit $i$ are defined as the real value of outcome for a unit $i$ with treatment value 
 $t=1$ and $t=0$,\footnote{Outcomes with $1-t$ are called counterfactual outcomes~\cite{yao2021survey}.} respectively~\cite{yao2021survey}. Additionally, the ITE is defined as $\tau_i = \mathbb{E}[Y_i^{1}\vert X_i=\boldsymbol{x}_i]-\mathbb{E}[Y_i^{0}\vert X_i=\boldsymbol{x}_i]$~\cite{yao2021survey}.


\paragraph{ITE estimation under heterogeneous interference.} This study aims to estimate the ITE from observational heterogeneous graph data. The data can be denoted by $(\bold{X},\bold{T},\bold{Y},\bold{H})$, where $\bold{X}=\{\boldsymbol{x}_i\}^n_{i=1}$, $\bold{T}=\{t_i\}^n_{i=1}$, and $\bold{Y}=\{y_i\}^n_{i=1}$. We assume that there exists interference between units in heterogeneous graphs. In this case,  the outcome of a unit is not only influenced by its own treatments and covariates but also influenced by those of its neighbors~\cite{pmlr-v130-ma21c,rakesh2018linked}. 
In heterogeneous graphs, every unit can receive interference from its heterogeneous neighbors through multi-view edges, so the interference in heterogeneous graphs is referred to as heterogeneous interference. Such  heterogeneous interference contains two types of interference: \emph{same-view interference} and cross-view interference. The former is that interference occurs within the same views, and the latter happens when interference propagates across different views through   multi-view edges. 
To formalize the ITE under heterogeneous interference, we use $\boldsymbol{s}_i$ to denote a summary vector of $\bold{X}_{-i}$ and $\bold{T}_{-i}$ on heterogeneous graphs $\bold{H}$, where the subscript $-i$ denotes all other units except $i$. The potential outcomes of the unit $i$  in heterogeneous graphs, denoted by $y^1_i(\boldsymbol{s}_i)$ and $y^0_i(\boldsymbol{s}_i)$,  are real outcomes for the unit $i$ under  $\boldsymbol{s}_i$ and treatment value $t=1$ and $t = 0$, respectively.
Then, we define the ITE under heterogeneous interference as follows:
\begin{equation}\label{ite:spillover}
\begin{aligned}
&\tau_i=\mathbb{E}[Y_i^{1}(S_i=\boldsymbol{s}_i)\vert X_i=\boldsymbol{x}_i]-\mathbb{E}[Y_i^{0}(S_i=\boldsymbol{s}_i)\vert X_i=\boldsymbol{x}_i].
\end{aligned}
\end{equation}

\paragraph{Confounder.} The existence of confounders is a well-known issue when estimating the ITE from observational data~\cite{pmlr-v70-shalit17a}. Confounders are parts of covariates, which can simultaneously  affect the treatment assignment and outcome~\cite{yao2021survey}, resulting in an imbalance in the distributions of different treatment assignments.
For instance, we consider that the treatment is whether a product is advertised. 
Famous brands have more promotion funds to advertise their products.
Meanwhile, customers tend to buy a product (e.g., a computer) from a famous brand (e.g., Apple).
In this case, the brand is a confounder. Without accurately addressing  confounders,  ITE estimation will be biased.


\paragraph{Assumption 1.} 
Following the previous studies~\cite{ma2022learning,pmlr-v130-ma21c}, we assume that there exists an aggregation function that can aggregate information of other units on heterogeneous graphs while outputting a vector $\boldsymbol{s}$, i.e., $\boldsymbol{s}_i=\text{AGG}(\bold{T}_{-i},\bold{X}_{-i},\bold{H})$. Here, we extend the neighbor interference assumption~\cite{doi:10.1080/01621459.2020.1768100} to heterogeneous interference,
for $\forall i$, $\forall \bold{T}_{-i},\bold{T}'_{-i},\forall \bold{X}_{-i},\bold{X}'_{-i}$, and $\forall \bold{H},\bold{H}'$: 
when $\boldsymbol{s}_i={\rm{AGG}}(\bold{T}_{-i},\bold{X}_{-i},\bold{H}) = {\rm{AGG}}(\bold{T}'_{-i},\bold{X}'_{-i},\bold{H}') = \boldsymbol{s}'_i $,  $Y^t_i(S_i=\boldsymbol{s}_i) = Y^t_i(S_i=\boldsymbol{s}'_i)$ holds.
 
\paragraph{Assumption 2.}  We extend consistency assumption~\cite{doi:10.1080/01621459.2020.1768100} to heterogeneous interference setting. We assume $Y_i=Y_i^{t_i}(S_i=\boldsymbol{s}_i)$ on the heterogeneous graphs $\bold{H}$ for 
 the unit $i$ with $t_i$ and $\boldsymbol{s}_i$.

\paragraph{Assumption 3.} To address confounders, we extend the unconfoundedness assumption~\cite{doi:10.1080/01621459.2020.1768100,ma2022learning} to the heterogeneous interference setting. 
For any unit $i$, given the covariates, the treatment assignment and output of the aggregation function are independent of potential outcomes, i.e., $T_i,S_i \perp \!\!\! \perp Y_i^1(\boldsymbol{s}_i),Y_i^0(\boldsymbol{s}_i) \vert X_i$.


\paragraph{Theoretical analysis.} To model potential outcomes using observed data under heterogeneous interference, we prove the identifiability of the expected potential outcome $Y_i^{t}(\boldsymbol{s}_i)$ ($t=1$ or $t=0$) based on the above assumptions as follows:
\begin{equation*}
\begin{aligned}
 &\mathbb{E}[Y_i\vert X_i=\boldsymbol{x}_i,T_i=t,X_{-i} = \bold{X}_{-i},T_{-i} = \bold{T}_{-i},H=\bold{H}]\\
 = &\mathbb{E}[Y_i\vert X_i=\boldsymbol{x}_i,T_i=t,S_i=\boldsymbol{s}_i]\quad \quad \quad \quad \quad \quad \quad \quad \quad\quad \:\text{(\emph{Assumption 1})}\\
  = &\mathbb{E}[Y^t_i(\boldsymbol{s}_i)\vert X_i=\boldsymbol{x}_i,T_i=t,S_i=\boldsymbol{s}_i] \quad \quad\; \quad\quad \quad\quad \quad\quad\text{(\emph{Assumptions 1} and \emph{2})}\\
 = &\mathbb{E}[Y^t_i(\boldsymbol{s}_i)\vert X_i=\boldsymbol{x}_i] \quad \quad\quad\quad\quad \quad \quad \quad \quad\quad  \quad \quad  \quad \quad \quad \text{(\emph{Assumption 3)}} 
\end{aligned}
\end{equation*}
Based on the above proof, once we aggregate $\bold{X}_{-i}$ and $\bold{T}_{-i}$ on heterogeneous graphs $\bold{H}$ into $\boldsymbol{s}_i$,
we can estimate the potential outcomes $Y_i^{1}(\boldsymbol{s}_i)$ and $Y_i^{0}(\boldsymbol{s}_i)$. This enables us to estimate the ITE using Eq.~(\ref{ite:spillover}).



\section{Proposed Method: Individual Treatment Estimator Under Heterogeneous Interference}\label{sec:proposed}
\begin{figure*}
\centering
\includegraphics[width=1.0\textwidth]{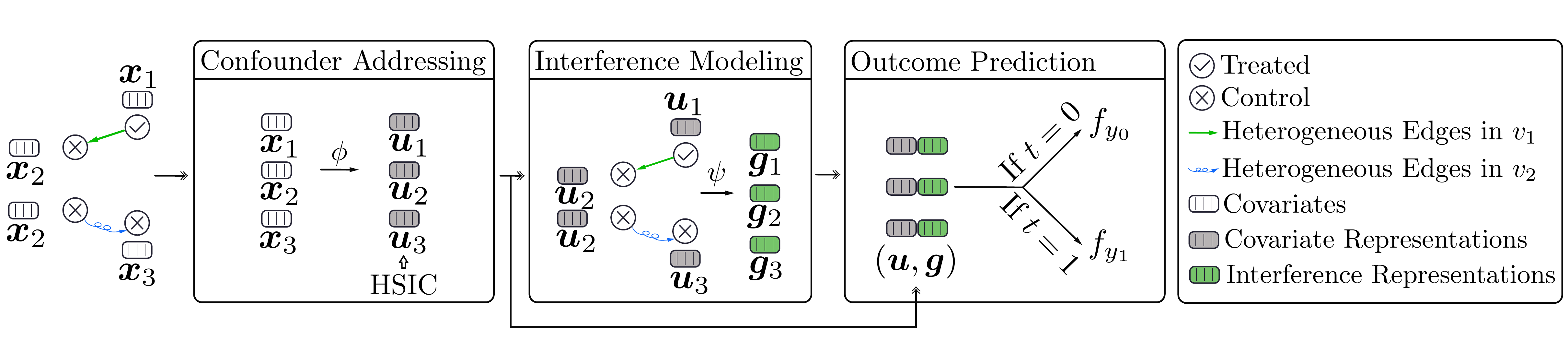}
\caption{An example of the model architecture of HINITE. In this case, there are two views, i.e., $v_1$ and $v_2$. 
}
\label{fig:HINITE}
\end{figure*}

This study proposes HINITE,  a method that can estimate the ITE from observed data $(\bold{X},\bold{T},\bold{Y},\bold{H})$ under heterogeneous interference. Figure~\ref{fig:HINITE} shows the architecture of HINITE. As can be seen, HINITE consists of three components to address confounders, model heterogeneous interference, and predict outcomes, respectively.  Specifically, the first component addresses confounders by learning balanced representations of covariates with the Hilbert-Schmidt Independence Criterion (HSIC) regularization~\cite{10.1007/11564089_7}.  The second component aggregates interference by modeling the propagation of interference  across units and views,  and generates representations of units, which are referred to as interference representations.  
The last component consists of two outcome predictors that infer potential outcomes using the covariate and interference representations.

\subsection{Learning Balanced Covariate Representations}\label{confounder} 

To address the imbalance in distributions of different treatment groups caused by confounders,  HINITE learns balanced covariate representations using an existing approach~\cite{pmlr-v130-ma21c}. The key idea is to find a representation space in which the treatment assignments and covariate representations become approximately independent~\cite{pmlr-v130-ma21c}.
This goal can be achieved by applying the HSIC regularization~\cite{10.1007/11564089_7}, which is an independence test criterion of two random variables. The value of HSIC is 0 when two random variables are independent. Thus, minimizing the HSIC can achieve the abovementioned goal. 

Specifically, we learn a balanced covariate representation $\boldsymbol{u}_i$ for the  $\boldsymbol{x}_i$ using a map function $\phi$ that consists of multiple feed-forward (FF) layers, i.e., $\boldsymbol{u}_i=\phi(\boldsymbol{x}_i)$, resulting in covariate representations for all units, denoted as $\bold{U}$. We train $\phi$ by minimizing the HSIC between $\boldsymbol{u}$ and $t$, which is denoted as $\rm{HSIC}_{\phi}$ and designed as follows:
\begin{equation}
    \mathrm{HSIC}_{\phi}(\bold{U},\bold{T})= \frac{1}{N^2}\mathrm{tr}(\bold{K}\bold{M}\bold{L}\bold{M}), \quad \bold{M} = \bold{I}_N - \frac{1}{N}\boldsymbol{1}_N\boldsymbol{1}_N^{\top},
\end{equation} 
where $N$ is the number of training units, $\cdot^{\top}$ represents the transposition operation, $\bold{I}_N$ is the identity matrix, and $\bold{1}_N$ is the vector of all ones.
$\bold{K}$ and $\bold{L}$ represent the Gaussian kernel applied to $\bold{U}$ and $\bold{T}$, respectively, i.e., 
\begin{equation}
    K_{ij}=\exp\left(-\frac{\Vert \boldsymbol{u}_i-\boldsymbol{u}_j\Vert^2_2}{2}\right),\quad L_{ij}=\exp\left(-\frac{(t_i-t_j)^{2}}{2}\right).
\end{equation}
\begin{figure}[htbp]
\centering
\includegraphics[width=1.0\textwidth]{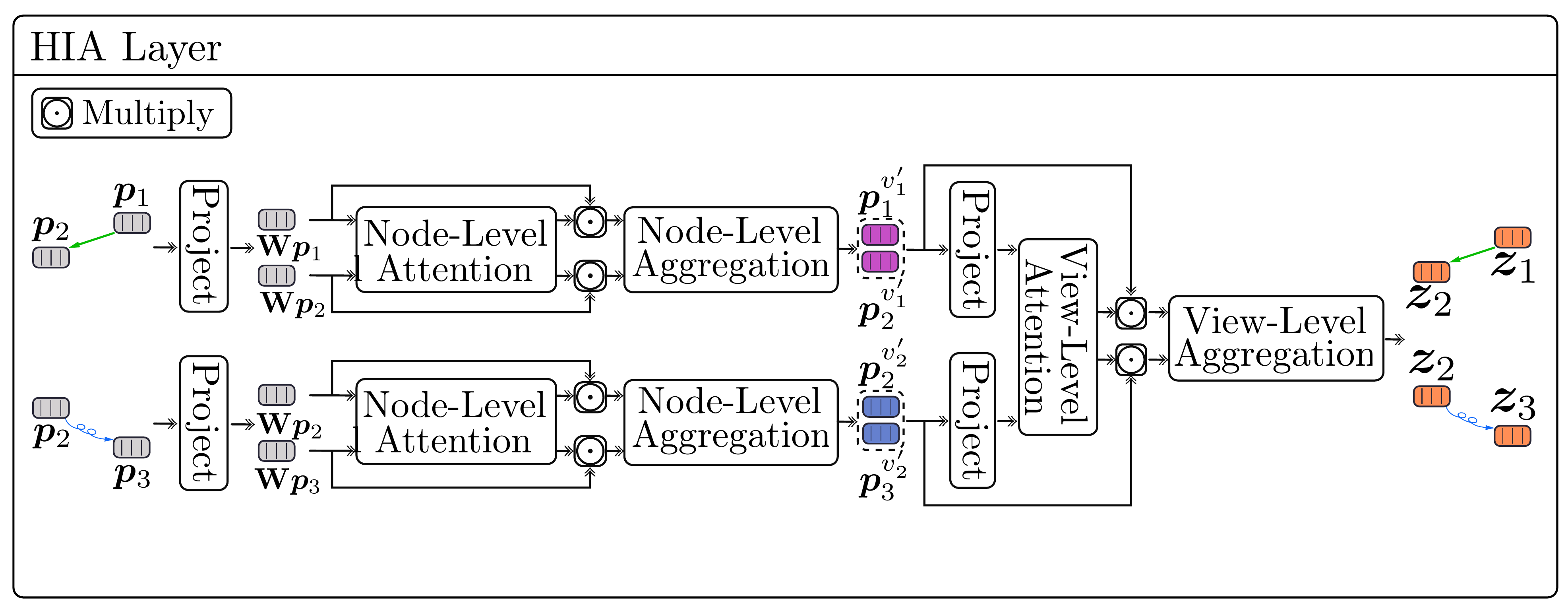}
\caption{The architecture of the HIA layer. This layer consists of node-level and view-level aggregation mechanisms with their attention mechanisms. }
\label{fig:hia}
\end{figure}



\subsection{Learning Heterogeneous Interference Representations}\label{interference}

To properly model heterogeneous interference, it is necessary to capture both same-view and cross-view interference.  To this end, we model the propagation of the same-view and cross-view interference. 
Inspired by Wang~\etal~\cite{wang2019heterogeneous}, we design an HIA layer, as shown in Figure~\ref{fig:hia}, which contains node-level and view-level aggregation mechanisms.
The node-level aggregation mechanism aggregates same-view interference received by units. It utilizes $m$ single-layered GNNs~\cite{kipf2016semi,vel2018graph} to perform aggregations within each view. The view-level aggregation mechanism combines (i.e., sums up) the results aggregated by the node-level aggregations to generate new representations of units. 
Therefore, by employing an HIA layer, units are able to aggregate interference received from their one-hop heterogeneous neighbors. This enables capturing cross-view interference by stacking HIA layers. Similarly, same-view interference from multi-hop neighbors can also be captured by stacking HIA layers.

Consider again the co-purchased and co-viewed graphs in Figure~\ref{fig:Interferenceb}.
Suppose that we feed units and their co-purchased and co-viewed graphs to a network stacked by two HIA layers. For the Mouse 1, the first HIA layer performs two node-level aggregations.  One aggregation helps the Mouse 1 aggregate interference within the co-purchased graph, while the other helps the Mouse 1 aggregate interference within the co-viewed graph, resulting in two aggregated results. Then, the view-level aggregation mechanism combines these  results obtained by node-level aggregations to generate the Mouse 1's new representation, while updating the new representation in all views. This enables the Mouse 1 to aggregate interference from the Computer.  Similarly, the first HIA layer also generates new representations for other units. Then, by taking these new representations of all units as inputs of the second HIA layer,  the second HIA layer enables the Mouse 2  to capture interference from the Mouse 1, which contains interference from the Computer. Therefore, the cross-view interference from the Computer to the Mouse 2 can be captured by stacking two HIA layers.

Apart from cross-view interference, another challenge is that the importance of edges  and  views may differ in heterogeneous graphs~\cite{wang2019heterogeneous}. For example,  in a co-viewed graph, the importance of products in the same category tends to be higher than that of products in different categories. Here, the weights of edges in the same view can be different. Furthermore, a co-purchased graph may have more significant importance than a co-viewed graph in terms of interference, leading to different importance for each view.  To overcome these difficulties and properly model the propagation of interference, we infer different weights for every edge via a graph attention mechanism~\cite{vel2018graph} (called node-level attention) before node-level aggregations, and learn different importance  for every view via an attention mechanism~\cite{vaswani2017attention,wang2019heterogeneous} (called view-level attention) before view-level aggregations.

More specifically, given covariate representations $\bold{U}$, treatment assignments $\bold{T}$, and structures of heterogeneous graphs $\bold{H}$, we aim to obtain interference representations $\bold{G}$ using a function $\psi$ that consists of multiple HIA layers, i.e., $\bold{G} =\psi(\bold{U},\bold{T},\bold{H})$. For a unit $i$, its  interference representation $\boldsymbol{g}_i$ is supposed to capture the interference from its heterogeneous neighbors.  Let $\boldsymbol{p}$ be a representation of a unit, which is the  input of the current HIA layer and the output of the previous HIA layer. For the first HIA layer,   
 $\boldsymbol{p}$ is the concatenation of $\boldsymbol{u}$ and $\boldsymbol{t}$.  Let $\boldsymbol{z}$ denote a new representation for the unit $i$ computed by the current HIA layer, $\alpha_{ij}^v$ denote the inferred weight of the edge between units $j$ and $i$ at the $v$-th view,  $w^v_i$ denote the learned importance of the $v$-th view for the unit $i$, and $\beta^v_i$ denote the normalized value for $w^v_i$.

Now, we describe the architecture of the HIA layer in detail. First, the HIA layer infers the edge weight $\alpha_{ij}^v$ by the node-level attention mechanism as follows:  
\begin{equation}
    \alpha_{ij}^v=\frac{\exp(\mathrm{LeakyReLU}(\boldsymbol{a}^{\top}[\bold{W}\boldsymbol{p}_i\Vert\bold{W}\boldsymbol{p}_j]))}{\sum_{k \in \bold{N}_{i}^v\bigcup{\{i\}}}\exp(\mathrm{LeakyReLU}(\boldsymbol{a}^{\top}[\bold{W}\boldsymbol{p}_i\Vert\bold{W}\boldsymbol{p}_k]))},
\end{equation}
where $\boldsymbol{a}$ and $\bold{W}$ represent a learnable parameter vector and matrix, respectively, and $\Vert$ represents the concatenation operation. 
Next, it performs node-level aggregations.  The node-level aggregation at the $v$-th view is computed as follows:
\begin{equation}
    \boldsymbol{p}^{v'}_{i}=\sigma\left(\sum_{j \in \bold{N}_{i}^v\bigcup{\{i\}}}\alpha_{ij}^v\bold{W}\boldsymbol{p}_j\right),
\end{equation}
where $\sigma$ is an activation function, such as ReLU.
Next,  the view-attention mechanism is applied to learn the importance of different views as follows:
\begin{equation}
    w^v_i = \frac{1}{n}\sum_{i=1}^{n}\boldsymbol{q}^{\top} \mathrm{LeakyReLU}{(\bold{W}\boldsymbol{p}^{v'}_{i} + \boldsymbol{b})}, \quad \beta^v_i = \frac{\exp{(w^v_i)}}{\sum_{v=1}^m\exp{(w^v_i)}},
\end{equation}
where $\boldsymbol{b}$ is a bias vector, and $\boldsymbol{q}$ is a learnable parameter vector. 
Finally, the view-level aggregation is applied to aggregate the information from different views as follows:
\begin{equation}
    \boldsymbol{z}_i = \sum_{v=1}^m \beta^v_i\boldsymbol{p}^{v'}_{i}.
\end{equation}

\subsection{Outcome Predictions and ITE estimation}
Given the covariate representations $\bold{U}$,  interference representations $\bold{G}$, and treatment assignments $\bold{T}$, we train two predictors that consist of multiple FF layers to infer the outcomes with different $t$.
Specifically, let $f_{y_0}$ and $f_{y_1}$  denote the predictor for $t=0$ and $t=1$, respectively. We optimize the two predictors by minimizing the following mean square error (MSE) between prediction outcomes and observed outcomes with the HSIC regularization:
\begin{equation}
\mathcal{L} = \frac{1}{N}\sum^{N}_{i=1}\left(f_{y_{t_i}}(\boldsymbol{u}_i,\boldsymbol{g}_i)-y_{i}\right)^2 +  \gamma \rm{HSIC}_{\phi},
\end{equation}
where the $\gamma$ is a regularization hyperparameter.

Finally, we can estimate the ITE using 
$
    \hat{\tau}_i=f_{y_1}(\boldsymbol{u}_i,\boldsymbol{g}_i)-f_{y_0}(\boldsymbol{u}_i,\boldsymbol{g}_i).
$

\section{Experiments}
\subsection{Datasets}
We used three heterogeneous graph datasets: Amazon Software (AMZ S)~\cite{he2016ups}, Youtube~\cite{tang2009uncoverning},  and 
 Flicker~\cite{wang2013learning}. Following prior studies on ITE/ATE~\cite{pmlr-v130-ma21c,ma2022learning,pmlr-v70-shalit17a}, we simulated outcomes\footnote{The simulated outcomes and the codes of the HINITE are available at \url{https://github.com/LINXF208/HINITE}.} as the ground-truth values for counterfactual outcomes are not available.
 
\noindent\textbf{Outcome Simulation:}
Similar to the outcome simulation in Ma~\etal~\cite{ma2022learning}, we used available data and heterogeneous graph structures to simulate outcomes under heterogeneous interference of the unit $i$:
\begin{equation} \label{sim:outcome1}
y_i = f_{0}(\boldsymbol{x}_i) + f_{t}(t_i,\boldsymbol{x}_i) + f_s(\bold{T},\bold{X},\bold{N}_i) + \epsilon_i,
\end{equation}
where $f_{0}(\boldsymbol{x}_i)=\boldsymbol{w}_0^{\top}\boldsymbol{x}_i$ simulates the outcome of a unit $i$ under treatment $t_i=0$ without interference, and every element of $\boldsymbol{w}_0$ follows a Gaussian distribution or uniform distribution (i.e., $\mathcal{N}(0,1)$ or $\mathcal{U}(0,1)$). $f_{t}(t_i,\boldsymbol{x}_i)=t_i\times\boldsymbol{w}_1^{\top}\boldsymbol{x}_i$ simulates the ITE of the unit $i$, where $\boldsymbol{w}_1\sim \mathcal{N}(0,\bold{I})$ or $\mathcal{U}(0,\bold{I})$. In the literature, the effect caused by interference is known as \emph{spillover effect}~\cite{rakesh2018linked}. We simulate it through  $f_s(\bold{T},\bold{X},\bold{N}_i)= o^{(1)}_i + o^{(2)}_i$, where $o^{(1)}_i=\text{Agg}(\text{Concat}(\bold{X},\bold{T}),\bold{N}_i)$ represents a spillover effect from 1-hop heterogeneous neighbors for the unit $i$, $o_i^{(2)}=\text{Agg}(\bold{O^{(1)}},\bold{N}_i)$ represents the spillover effect of 2-hop heterogeneous neighbors, and $\bold{O^{(1)}}$ represents the spillover effects from 1-hop heterogeneous neighbors for all units. Here, the aggregation function is defined as $\text{Agg}(\bold{C},\bold{N}_i)= \sum_{v=1}^m e^v\left(\frac{1}{|\bold{N}_{i}^v|}\sum_{j \in \bold{N}_{i}^v}\boldsymbol{w}_{ij}^{\top}\boldsymbol{c}_j\right)$,
where $e^{v}$ and every element of $\boldsymbol{w}_{ij}$ follow $\mathcal{N}(0,1)$ or $\mathcal{U}(0,1)$. Lastly, $\epsilon_i\sim \mathcal{N}(0,1)$ is a random noise. 

\noindent\textbf{Amazon Software dataset~\cite{he2016ups}:} The Amazon dataset~\cite{he2016ups} is collected from Amazon\footnote{\url{https://www.amazon.com/}}. In the graphs of the Amazon dataset, each node is a product.
To study causal effects, we chose the co-purchased and co-viewed graphs from the software category of the Amazon dataset. 
After removing nodes with missing values, there are 11,089 items with 11,813 heterogeneous edges. The covariates consist of reviews and the number of customer reviews of items. We put reviews into the SimCSE~\cite{gao2021simcse} model to generate 768-dimensional sentence embeddings. The review rating of items is considered as a treatment: an item is treated ($t=1$) when the average review rating is at least 3, and an item is controlled ($t=0$) when the average review rating is less than 3. 
The causal problem in this dataset is whether review rating has a role in influencing the sales of items. Due to the heterogeneous edges among items, the sales of an item might be influenced by its heterogeneous neighbors' treatments.

\noindent\textbf{YouTube dataset~\cite{tang2009uncoverning}:} 
 Tang~\etal~\cite{tang2009uncoverning} used YouTube Data
API\footnote{\url{https://developers.google.com/youtube/?csw=1}} to crawl the information of contacts, subscriptions, and favorites of users from YouTube\footnote{\url{https://www.youtube.com/}}, while extending them to a contact graph, co-subscription graph,  co-subscribed graph, and favorite graph. Every node in the graphs is a user of YouTube. In this case, we consider a causal problem:  “how much recommendation of a video (treatment) to a user will affect the user's experience of this video (outcome)?” Moreover, users might share the recommended video with heterogeneous neighbors, which constitutes heterogeneous interference. We took 5,000 users with their heterogeneous graphs containing 3,190,622 heterogeneous edges to simulate outcomes and study heterogeneous interference.
As detailed information about each user is missing, we simulated the covariates via $\boldsymbol{x}_i\sim\mathcal{N}(0,\bold{I})$ (100-dimensional vector), and simulated treatment $t_i$ as follows,  following most existing works, such as Ma~\etal~\cite{ma2022learning}:
\begin{equation} \label{sim:t}
t_i\sim \text{Ber}(\text{sigmoid}(\boldsymbol{x}_i^{\top}\boldsymbol{w}_t)+\epsilon_{t_i}),
\end{equation}
where $\text{Ber}(\cdot)$ represents a Bernoulli distribution, $\boldsymbol{w}_t$ is a 100-dimensional vector in which every element follows $\mathcal{U}(-1,1)$, and $\epsilon_{t_i}$ is random Gaussian noise.

\noindent\textbf{Flicker dataset~\cite{wang2013learning}:} 
 Flicker\footnote{\url{https://www.flickr.com/}} is an online social website where users can share their images. Qu~\etal~\cite{qu2017attention} constructed a dataset with multi-view graphs, i.e., friendship view and similarity view, from the Flicker dataset~\cite{wang2013learning}. 
 Every node in the graphs is a user of Flicker. Following Qu~\etal~\cite{qu2017attention}, we also consider friendship-view and similarity-view graphs that have 7,575 users with approximately 1,236,976 heterogeneous edges.
Here, the causal question is: “how much  recommending a hot photo (treatment) to a user will affect the user's experience  (outcome) of this photo?” In this case, users might share recommended photos with their heterogeneous neighbors, which constitutes heterogeneous interference.
We used the 1206-dimensional embeddings that are provided by Guo~\etal~\cite{guo2020learning},  generated using a list of users' interest tags, and simulated the treatments using Eq.~(\ref{sim:t}).

\subsection{Baselines}
\noindent\textbf{BNN~\cite{Johansson2016}:} Balancing Neural Network~\cite{Johansson2016} (BNN) addresses confounders by minimizing the discrepancy of distributions of units belonging to different groups, without considering interference. Following Johansson~\etal~\cite{Johansson2016}, we considered two structures: BNN-4-0 and BNN-2-2. The former has four representation layers but no prediction layers, and the latter has two representation layers and two prediction layers. Both have one linear output layer. 
 
\noindent\textbf{CFR~\cite{pmlr-v70-shalit17a}:} Counterfactual Regression (CFR)~\cite{pmlr-v70-shalit17a} minimizes the maximum mean discrepancy (MMD) and Wasserstein distance between different distributions of two groups. Similar to BNN, it also ignores interference. Following Shalit \etal~\cite{pmlr-v70-shalit17a}, we considered two different schemes:  CFR$_{\rm{MMD}}$ and  CFR$_{\rm{Wass}}$. The former minimizes the MMD of two different distributions, while the latter minimizes the Wasserstein distance.

\noindent\textbf{TARNet~\cite{pmlr-v70-shalit17a}:} TARNet consists of the same model architecture as the CFR model but removes the balance term (MMD or Wasserstein distance).

\noindent\textbf{GCN-based methods~\cite{pmlr-v130-ma21c}:} Ma~\etal~\cite{pmlr-v130-ma21c} proposed methods to address interference on a homogeneous graph using graph convolutional networks (GCNs)~\cite{https://doi.org/10.48550/arxiv.1609.02907}.  The GCN-based method can use only a single view rather than all views of heterogeneous graphs. To overcome it,  we consider two schemes. The first scheme is to replace heterogeneous graphs with a projection graph $\mathbf{A}_{\rm{Proj}}$ and apply the GCN-based method to the $\mathbf{A}_{\rm{Proj}}$, denoted as $\rm{GCN}_{\rm{Proj}}$. If two units have an edge in either of the original heterogeneous graphs, there will be an edge in this projection graph. The second scheme is to augment the GCN-based method with mixing operations, which includes two variants: MGCN$_{\rm{C}}$ and MGCN$_{\rm{M}}$. The MGCN$_{\rm{C}}$ concatenates interference representations from different views into a single vector, while the MGCN$_{\rm{M}}$ computes the mean vector of these interference representations. 



\subsection{Experiment Settings}

For all datasets, we calculated  $\epsilon_{\rm{PEHE}}$/$\epsilon_{\rm{ATE}}$ to evaluate the error on ITE/ATE estimations as follows:
\begin{equation}
    \epsilon_{\mathrm{PEHE}}=\frac{1}{n}\sum^n_{i=1}(\tau_i-\hat{\tau}_i)^2,\quad \epsilon_{\mathrm{ATE}}=\left\vert\frac{1}{n}\sum_{i =1}^{n}\tau_i-\frac{1}{n}\sum_{i =1}^{n}\hat{\tau}_i\right\vert.
\end{equation}

Following Ma~\etal~\cite{pmlr-v130-ma21c}, the entire $\bold{X}$, $\bold{T}$, and heterogeneous graph structures were given during the training, validation, and testing phases. However, only the observed outcomes of the units in the training set were provided during the training phase. 
\begin{table}[]
\centering
\caption{Results (mean ± standard errors) of performance of ITE and ATE estimation. Results in bold indicate the lowest mean error. HINITE is our method. 
}
\resizebox{\textwidth}{!}{ 
\begin{tabular}{lcccccc}
\toprule
& \multicolumn{2}{c}{Youtube}     & \multicolumn{2}{c}{Flicker}        &  \multicolumn{2}{c}{AMZ S}             \\
Method & $\epsilon_{\textrm{PEHE}}$ & $\epsilon_{\textrm{ATE}}$ & $\epsilon_{\textrm{PEHE}}$ & $\epsilon_{\textrm{ATE}}$ & $\sqrt{\epsilon_{\textrm{PEHE}}}$ & $\epsilon_{\textrm{ATE}}$  \\ \midrule
TARNet           						& 40.75$\pm$7.95       & 0.51$\pm$0.23  & 24.20$\pm$6.79   & 0.30$\pm$0.26  & 112.37$\pm$11.54  & 103.91$\pm$12.78\enspace\\ 
 BNN-2-2           						&    \enspace93.03$\pm$16.02        & $0.26$$\pm$$0.23$  & 27.91$\pm$7.53  & $\textbf{0.13}$$\pm$$\textbf{0.07}$  & 199.37$\pm$0.20\enspace      & 196.36$\pm$0.20\enspace\enspace\\
BNN-4-0          				& 105.38$\pm$22.50   & 0.26$\pm$0.23		& 29.22$\pm$7.53        & \textbf{0.13}$\pm$\textbf{0.07}  & 206.03$\pm$0.08\enspace       & 203.12$\pm$0.08\enspace\enspace \\
CFR$_{\rm{MMD}}$    	& 42.02$\pm$9.96       & 0.43$\pm$0.36  & 24.44$\pm$7.49  & 0.29$\pm$0.17 & 103.18$\pm$25.02   & 89.76$\pm$32.13 \\
CFR$_{\rm{WASS}}$     &  39.36$\pm$8.76   & 0.51$\pm$0.41 &24.02$\pm$6.71      & 0.35$\pm$0.17  &  109.91$\pm$24.49   & 99.40$\pm$30.40\\ 
GCN$_{\text{Proj}}$ 	& 42.37$\pm$7.45        & 0.61$\pm$0.39  & 24.59$\pm$5.11   & 0.21$\pm$0.13 & 139.14$\pm$20.63   & 135.57$\pm$22.86\enspace  \\
MGCN$_{\rm{C}}$  	& \enspace53.10$\pm$11.83        & 0.29$\pm$0.27  & 26.87$\pm$6.43   & 0.25$\pm$0.20 & 95.14$\pm$8.25   & 72.08$\pm$13.47  \\
MGCN$_{\rm{M}}$  	& \enspace53.99$\pm$13.46       & 0.37$\pm$0.33  & 29.48$\pm$7.17   & 0.29$\pm$0.25 & 87.33$\pm$3.40   & 60.81$\pm$3.27\enspace  \\\midrule
HINITE & 	\textbf{14.43}$\pm$\textbf{3.27}        & \textbf{0.21}$\pm$\textbf{0.20}  & \textbf{18.45}$\pm$\textbf{4.42}   & 0.15$\pm$0.11  & \textbf{76.16}$\pm$\textbf{3.82}   & \textbf{15.21}$\pm$\textbf{3.89}\enspace      \\

\midrule
\end{tabular}
}
\label{tab:ITE_performance}
\end{table}
We randomly split all datasets into training/validation/test splits  with a ratio of 70\%/15\%/15\%. Results on the Youtube and Flicker datasets were averaged over ten realizations, while the results on the AMZ S dataset were averaged over three repeated executions. We trained all models with the NVIDIA RTX A5000 GPU. All methods utilized the Adam optimizer with 2,000 training iterations for all datasets. In addition, dropout and early stopping were applied for all methods to avoid overfitting.


For all datasets, we set the learning rate to 0.001 with a weight decay of 0.001, set the training batch size to 512, and searched $\gamma$ in the range of $\{0.01,0.1,0.5,1.0,1.5\}$ using the validation sets. We used ReLU as activation function for $\phi$, $f_{y_{t_i}}$, and node-level aggregations. The hidden layers of $\phi$ were set to $(128,64,64)$-dimensions,  $\psi$ are set to $(64,64,32)$-dimensions,  $f_{y_{t_i}}$ are set to $(128,64,32)$-dimensions, and the dimensions of view-level attention were set to $(128,128,64)$-dimensions.  Moreover, we  searched for hyperparameters for all baseline methods from the search range suggested in the corresponding literature.


\subsection{Results}

\paragraph{Treatment effect estimation performance.}Table~\ref{tab:ITE_performance} lists the results of ITE and ATE estimations on test sets of all datasets. It can be seen that  the HINITE outperforms all baseline methods in ITE estimation, while there are significant gaps (p-values of the t-test are far less than 0.05) in ITE estimation between the proposed and baseline methods. It can also be seen that  HINITE outperforms most baseline methods in ATE estimation, at least, achieving comparative performance of ATE estimation to those of the baseline methods. These results reveal that HINITE has a powerful ability to address heterogeneous interference. Moreover, the GCN$_{\rm{Proj}}$ and MGCN with some simple mixers cannot always achieve better performance than other baseline methods. This implies that modeling cross-view interference using the HIA layers is important. 

\begin{figure}
	\centering
	\subfloat[Flicker, $\epsilon_{\textrm{PEHE}}$]{\includegraphics[width=0.45\columnwidth]{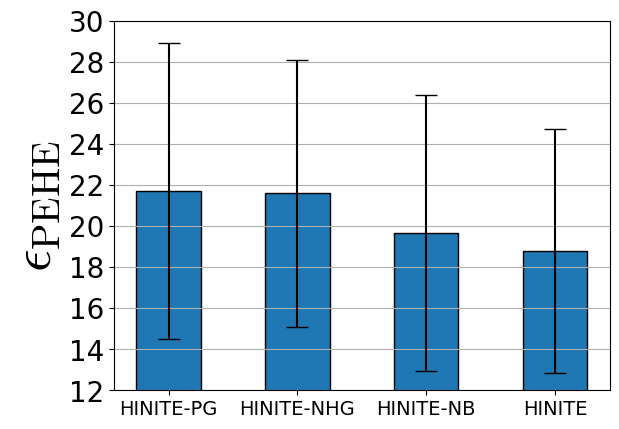}}\hspace{3pt}
	\subfloat[Flicker, $\epsilon_{\textrm{ATE}}$]{\includegraphics[width=0.45\columnwidth]{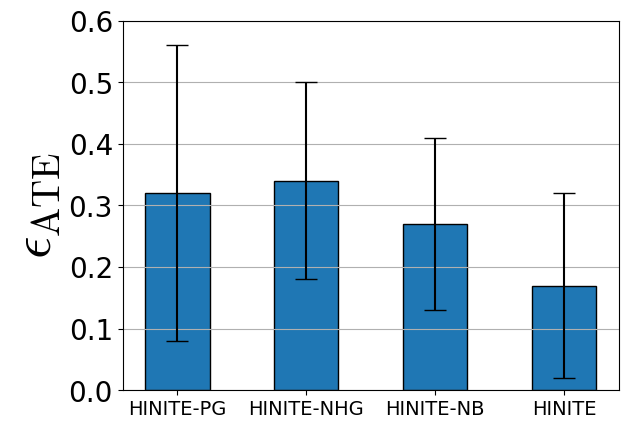}}\\
\caption{Results (mean and standard error) of ablation experiments. We set $\alpha$ to 1.5 for HINITE-PG, HINITE-NHG, and HINITE in the ablation experiments. }
\label{fig:ab}	
\end{figure}
\begin{figure}[htbp]
	\centering
	\subfloat[Flicker, $\epsilon_{\textrm{PEHE}}$]{\includegraphics[width=.45\columnwidth]{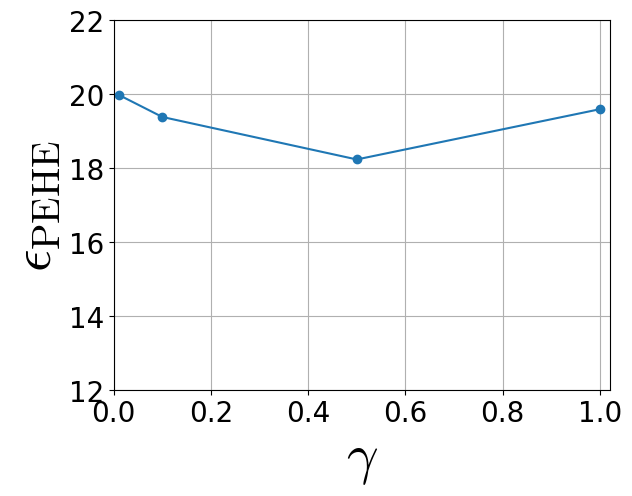}} \hspace{3pt}
	\subfloat[Flicker, $\epsilon_{\textrm{ATE}}$]
 {\includegraphics[width=.45\columnwidth]{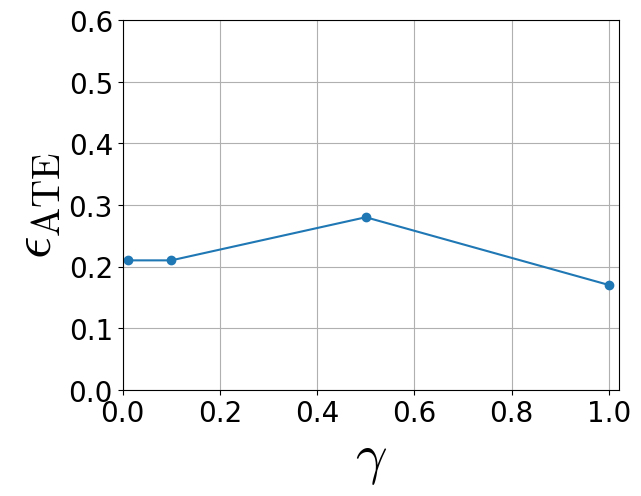}}
 
	\subfloat[Youtube, $\epsilon_{\textrm{PEHE}}$]{\includegraphics[width=.45\columnwidth]{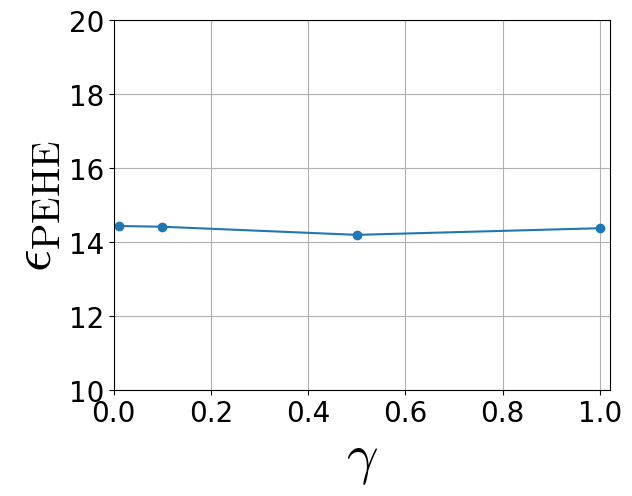}}  \hspace{3pt}
	\subfloat[Youtube, $\epsilon_{\textrm{ATE}}$]{\includegraphics[width=.45\columnwidth]{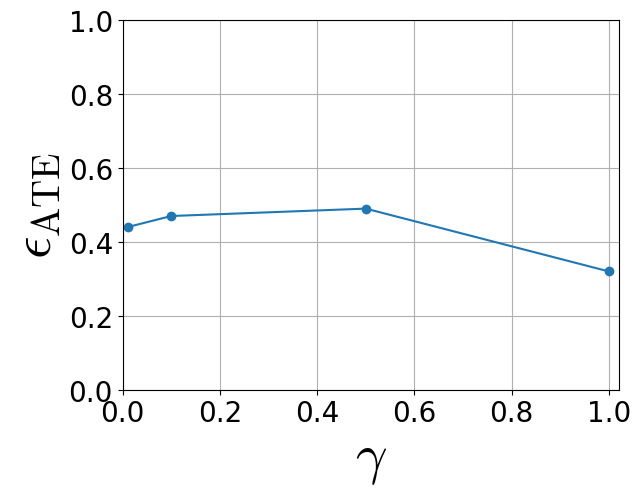}}
\caption{Performance changes on Filker and Youtube datasets with different $\gamma$ (in the range of $\{0.01, 0.1, 0.5, 1.0\}$) . Results are averaged over ten realizations with a fixed value of $\gamma$.}
\label{fig:sens}
\end{figure}

\paragraph{Ablation study.} To further investigate the importance of each component of HINITE, we conducted ablation experiments. Let us start by introducing some  variants of HINITE: (i) 
HINITE-PG applies the HINITE to the projection graph $\mathbf{A}_{\rm{Proj}}$, which was described when introduced  the GCN-based method.
(ii) HINITE-NHG replaces the HIA layers with GCN layers~\cite{kipf2016semi} while using the $\mathbf{A}_{\rm{Proj}}$. (iii) HINITE-NB removes the HSIC regularization 
by setting $\gamma$ to 0.  

Figure~\ref{fig:ab} presents the results of the ablation experiments. A clear performance gap can be seen in ITE and ATE estimation between the HINITE-PG/HINITE-NHG and HINITE. This implies that it is important to model the heterogeneous interference using the information of heterogeneous graphs and the proposed HIA layer. Comparing the results of HINITE and HINITE-NB, we can also observe that removing the HSIC regularization results in performance degradation.  This reveals that it is also important to balance the different distributions.
\paragraph{Sensitivity analysis.} To investigate whether HINITE is sensitive to $\gamma$, we conducted experiments with different $\gamma$ and present the results in Figure~\ref{fig:sens}. No significant change in performance was observed with different values of $\gamma$. This reveals that HINITE is not particularly sensitive to the value of  $\gamma$.
\section{Related work}
In the literature, efforts have been made to estimate treatment effect without interference~\cite{chu2021graph,guo2020learning,Johansson2016,li2022deep,10.1093/biomet/70.1.41,rubin1980randomization,pmlr-v70-shalit17a,yao2021survey,NEURIPS2018_a50abba8} and with interference on homogeneous graphs ~\cite{Aronow2017EstimatingAC,doi:10.1080/01621459.2020.1768100,doi:10.1198/016214508000000292,liu2014large,pmlr-v130-ma21c,tchetgen2012causal,tchetgen2021auto,RePEc:arx:papers:1906.10258} or hyper-graphs~\cite{ma2022learning}. A few studies have considered heterogeneous graphs.
For example, Qu~\etal~\cite{qu2021efficient} assumed a partial interference and could only estimate ATE.
Zhao~\etal~\cite{zhao2022learning} proposed a method to construct a heterogeneous graph from a homogeneous graph by learning a set of weights for each edge using an attention mechanism, but their method cannot capture interference between multi-view graph structures. We offer the first approach for handling interference on multi-view graphs.

Meanwhile,  heterogeneous graphs have been the subject of recent graph analysis studies, focusing on tasks such as node classification, link prediction, and graph classification~\cite{fu2020magnn,jin2021heterogeneous,liang2022meta,shi2021entity,song2022self,wang2022survey,wang2019heterogeneous,zhang2019heterogeneous,zhao2021heterogeneous}. 
The proposed HINITE shares some similarities with the heterogeneous graph attention network (HAN)~\cite{wang2019heterogeneous}. However, HAN aggregates information from each view at the end of forward propagation only once, while the proposed HINITE does aggregation layer-by-layer, which is essential for capturing cross-view interference. In addition, we use LeakyReLU (for view-level attention) instead of the tanh function as an activation function to address the vanishing gradient issue, and we use single-head instead of multi-head attention for better efficiency.

\section{Conclusion}

In this paper, we described the problem of heterogeneous interference and the difficulty of treatment effect estimations under heterogeneous interference. This paper proposed HINITE to model the propagation of heterogeneous interference using HIA layers that contain node-level aggregation, view-level aggregation, and attention mechanisms. We conducted extensive experiments to verify the performance of the proposed HINITE, where the results validate the effectiveness of the HINITE in ITE and ATE estimation under heterogeneous interference.

\section*{Acknowledgements}
This work was supported by JST, the establishment of university fellowships towards the creation of science technology innovation, Grant Number JPMJFS2123, and supported by JSPS KAKENHI Grant Number 20H04244.

\clearpage
\section*{Ethics}
This study only involved public datasets that are freely available for academic purposes.

\bibliographystyle{splncs04}
\bibliography{main}

\clearpage

\end{document}